\documentclass{article}

\usepackage{microtype}
\usepackage{graphicx}
\usepackage{subfigure}
\usepackage{booktabs} 

\usepackage{hyperref}       
\usepackage{url}            
\usepackage{booktabs}       
\usepackage{amsfonts}       
\usepackage{nicefrac}       
\usepackage{microtype}      
\usepackage{xcolor}         
\usepackage{graphicx}
\usepackage{float} 
\usepackage{subfigure}
\usepackage{amsmath}
\usepackage[ruled,vlined]{algorithm2e}
\usepackage{algorithmic}
\usepackage{color, colortbl}
\usepackage{wrapfig}
\usepackage{ulem}
\usepackage{multirow}
\makeatletter

\newcommand{\Rmnum}[1]{\expandafter@slowromancap\romannumeral #1@}
\makeatother
\usepackage{caption}
\definecolor{light-gray}{gray}{0.90}

\usepackage[accepted]{icml2023}

\usepackage[capitalize,noabbrev]{cleveref}

\usepackage[textsize=tiny]{todonotes}
\icmltitlerunning{Why Deep Models Often Cannot Beat Non-deep Counterparts on Molecular Property Prediction?}

\begin{document}

\twocolumn[
\icmltitle{Why Deep Models Often Cannot Beat Non-deep Counterparts \\ on Molecular Property Prediction?}



\icmlsetsymbol{equal}{*}

\begin{icmlauthorlist}
\icmlauthor{Jun Xia}{equal,yyy}
\icmlauthor{Lecheng Zhang}{equal,yyy}
\icmlauthor{Xiao Zhu}{equal,yyy}
\icmlauthor{Stan Z. Li}{yyy}
\end{icmlauthorlist}

\icmlaffiliation{yyy}{Westlake University, Hangzhou, China}

\icmlcorrespondingauthor{Jun Xia}{xiajun@westlake.edu.cn}
\icmlcorrespondingauthor{Stan Z. Li}{stan.zq.li@westlake.edu.cn}

\icmlkeywords{Machine Learning, ICML}

\vskip 0.3in
]



\printAffiliationsAndNotice{\icmlEqualContribution} 

\begin{abstract}
Molecular property prediction (MPP) is a crucial task in the drug discovery pipeline, which has recently gained considerable attention thanks to advances in deep neural networks. However, recent research has revealed that deep models struggle to beat traditional non-deep ones on MPP. In this study, we benchmark 12 representative models (3 non-deep models and 9 deep models) on 14 molecule datasets. Through the most comprehensive study to date, we make the following key observations: \textbf{(\romannumeral 1)} Deep models are generally unable to outperform non-deep ones; \textbf{(\romannumeral 2)} The failure of deep models on MPP cannot be solely attributed to the small size of molecular datasets. What matters is the irregular molecule data pattern; \textbf{(\romannumeral 3)} In particular, tree models using molecular fingerprints as inputs tend to perform better than other competitors.
Furthermore, we conduct extensive empirical investigations into the unique patterns of molecule data and inductive biases of various models underlying these phenomena.

\end{abstract}
\vspace{-1.8em}
\section{Introduction}
\begin{figure*}[t]
    \begin{center}
    \includegraphics[width=0.99\linewidth]{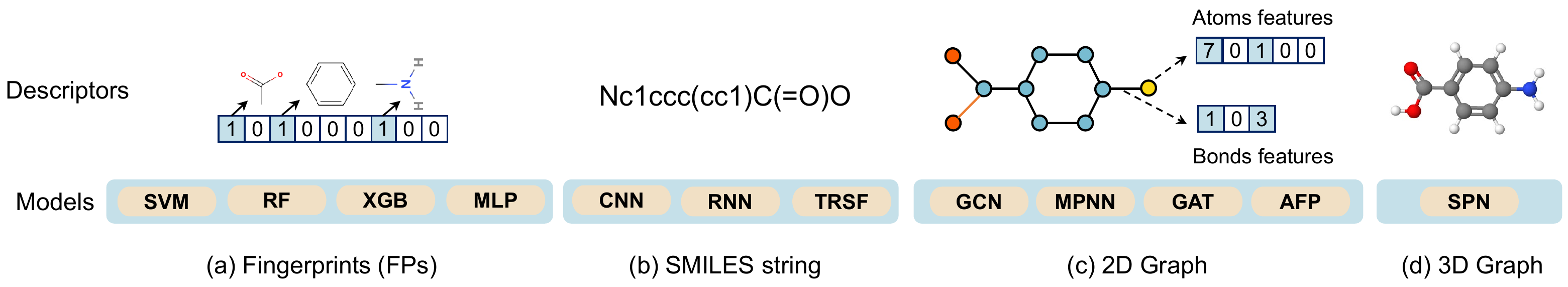}
    \end{center}
    \vspace{-1em}
    \caption{Exemplary molecular descriptors and their corresponding models in our benchmark. \textbf{SVM}: Support Vector Machine~\cite{zernov2003drug}; \textbf{RF}: Random Forest~\cite{svetnik2003random}; \textbf{XGB}: eXtreme Gradient Boosting~\cite{chen2016xgboost}; \textbf{MLP}: Multi-Layer Perceptron; \textbf{CNN}: 1D Convolution Neural Network~\cite{kimber2021maxsmi}; \textbf{RNN}: Recurrent Neural Network (GRU)~\cite{Mulder2015ASO}; \textbf{TRSF}: TRanSFormer~\cite{vaswani2017attention}; \textbf{GCN}: Graph Convolution Network~\cite{kipf2017semi-supervised}; \textbf{MPNN}: Message-Passing Neural Network~\cite{gilmer2017neural}; \textbf{GAT}: Graph Attention neTwork~\cite{velickovic2018graph}; \textbf{AFP}: Attentive FP~\cite{xiong2020pushing}; \textbf{SPN}: SPhereNet~\cite{liu2022spherical}. The above-mentioned abbreviations are applicable throughout the entire paper.}
     \label{fig1}
     \vspace{-1em}
\end{figure*}

Molecular Property Prediction (MPP) is a critical task in drug discovery, aimed at identifying molecules with desirable pharmacological and ADMET (absorption, distribution, metabolism, excretion, and toxicity) properties. Machine learning models have been widely used in this fast-growing field, with two types of models being commonly employed: traditional non-deep models and deep models. In non-deep models, molecules are fed into traditional machine learning models such as Random Forest and Support Vector Machine in the format of computed or handcrafted molecular fingerprints~\cite{todeschini2010molecular}. The other group utilizes deep models to extract expressive representations for molecules in a data-driven manner. Specifically, the Multi-Layer Perceptron (MLP) could be applied to computed or handcrafted molecular fingerprints; Sequence-based neural architectures including Recurrent Neural Networks (RNNs)~\cite{medsker1999recurrent}, 1D Convolutional Neural Networks (1D CNNs)~\cite{gu2018recent}, and Transformers~\cite{honda2019smiles,rong2020self} are exploited to encode molecules represented in Simplified Molecular-Input Line-Entry System (SMILES) strings~\cite{weininger1989smiles.}. Later, it is argued that
molecules can be naturally represented in graph structures with atoms as nodes and bonds as edges. This inspires a line of works to leverage such structured inductive bias for better molecular representations~\cite{gilmer2017neural,xiong2019pushing,yang2019analyzing,song2020communicative}. The key advancements underneath these approaches are Graph Neural Networks (GNNs), which consider graph structures and attributive features simultaneously by recursively aggregating node features from neighborhoods~\cite{kipf2017semi-supervised,velickovic2018graph,hamilton2017inductive}. More recently, researchers incorporate 3D conformations of molecules into their representations for better performance, whereas pragmatic considerations such as calculation cost, alignment invariance, and uncertainty in conformation generation limited the practical applicability of these models~\cite{axen2017simple,Gasteiger2020Directional,schuett2017schnet,gasteiger2021gemnet,liu2022spherical}. We summarize the widely-used molecular descriptors and their corresponding models in our benchmark, as shown in Figure~\ref{fig1}.
Despite the fruitful progress, previous studies~\cite{mayr2018large,yang2019analyzing,valsecchi2022predicting,jiang2021could,van2022exposing,janela2022simple} have observed that deep models struggled to outperform non-deep ones on molecules. However, these studies neither consider the emerging powerful deep models (e.g., Transformer~\cite{honda2019smiles}, SphereNet~\cite{liu2021spherical}) nor explore various molecular descriptors (e.g., 3D molecular graph). Also, they did not investigate the reasons why deep models often fail on molecules.

To narrow this gap, we present the most comprehensive benchmark study on molecular property prediction to date, with a precise methodology for dataset inclusion and hyperparameter tuning. Our empirical results confirm the observations of previous studies, namely that deep models generally cannot outperform traditional non-deep counterparts. Moreover, we observe several interesting phenomena that challenge the prevailing beliefs of the community, which can guide optimal methodology design for future studies. Furthermore, we transform the original molecular data to observe the performance changes of various models, uncovering the unique patterns of molecular data and the differing inductive biases of various models. These in-depth empirical studies shed light on the benchmarking results.

\section{Benchmarking Results.}
In this section, we present a benchmark on 14 molecular datasets with 12 representative models. 

\begin{table*}[t]
\caption{The comparison of representative models on multiple molecular datasets. The standard deviations can be seen in the appendix for the limited space. \textbf{No.}: Number of the molecules in the datasets. The top-3 performances on each dataset are highlighted with the grey background. The best performance is highlighted with \textbf{bold}. Kindly note that \textbf{`TRSF'} denotes the transformer that has been pre-trained on 861, 000 molecular SMILES strings. The results on QM 9 can be seen in the appendix.}
\label{tab:main_results}
\setlength{\tabcolsep}{1.68pt}
\centering
\fontsize{7.8pt}{\baselineskip}\selectfont
\resizebox{0.68\textwidth}{!}{
\begin{tabular}{@{}cccccccccccccc@{}}
\toprule
\textbf{Dataset (No.)}  & \textbf{Metric} & \textbf{SVM} & \textbf{XGB} & \textbf{RF} & \textbf{CNN}& \textbf{RNN} & \textbf{TRSF}  & \textbf{MLP} & \textbf{GCN} & \textbf{MPNN} & \textbf{GAT} & \textbf{AFP} &\textbf{SPN}
\\ \midrule
BACE (1,513)& AUC\_ROC    &\cellcolor{light-gray}0.886  & \cellcolor{light-gray}\textbf{0.896}    & \cellcolor{light-gray}0.890  & 0.815 & 0.559    &0.835  & 0.887  & 0.880 &   0.846  &0.886   &  0.879  & 0.882    \\
HIV (40,748)& AUC\_ROC   & 0.817 &\cellcolor{light-gray}0.823   &\cellcolor{light-gray}0.826   & 0.733 & 0.639    &0.748  &0.791   &\cellcolor{light-gray}\textbf{0.834}  & 0.814    & 0.812  &  0.819     & 0.818 \\
BBBP (2,035)& AUC\_ROC    & 0.913 & \cellcolor{light-gray}\textbf{0.926}   & \cellcolor{light-gray}0.923  & 0.760 & 0.693    & 0.897 &\cellcolor{light-gray}0.918  & 0.915 &0.872     &0.902   &  0.893     &  0.905 \\
ClinTox (1,475) & AUC\_ROC     &0.879  & \cellcolor{light-gray}0.919   & \cellcolor{light-gray}0.933  &0.685  &  0.813   & \cellcolor{light-gray}\textbf{0.963} & 0.890  & 0.889 & 0.868    & 0.891  &0.907   & 0.912    \\
SIDER (1,366) & AUC\_ROC    & 0.626 &\cellcolor{light-gray}0.638   &\cellcolor{light-gray}\textbf{0.644}   &0.591  & 0.515    & \cellcolor{light-gray}0.641 & 0.617  &0.633  & 0.603    & 0.614  &  0.620  &  0.613   \\
Tox21 (7,811) & AUC\_ROC    & 0.820 & \cellcolor{light-gray}0.837    &\cellcolor{light-gray}0.838   & 0.766 &  0.734   & 0.817 &0.834   &0.830  & 0.816   & 0.829  &  \cellcolor{light-gray}\textbf{0.845}  & 0.827      \\
ToxCast (8,539)& AUC\_ROC    & 0.725 &\cellcolor{light-gray}0.785 & 0.778  & 0.735  & 0.74     &\cellcolor{light-gray} 0.780  & 0.781  &0.767  & 0.736    &0.768   & \cellcolor{light-gray}\textbf{0.788}    & 0.772   \\
MUV (93,087)& AUC\_PRC    &\cellcolor{light-gray}\textbf{0.093}  & \cellcolor{light-gray}0.072   & \cellcolor{light-gray}0.069  & 0.045  & 0.094    & 0.059 &0.018 & 0.056  & 0.019 & 0.055    & 0.044      &  0.058 \\
SARS-CoV-2 (14,332) & AUC\_ROC  &0.599  & \cellcolor{light-gray}\textbf{0.700}   & \cellcolor{light-gray}0.686  & \cellcolor{light-gray}0.688 & 0.649    & 0.643  & 0.638  & 0.646 & 0.640    & 0.683  &  0.651   & 0.663\\
\midrule
ESOL (1,127) & RMSE     & 0.676 &\cellcolor{light-gray}\textbf{0.583}    &\cellcolor{light-gray}0.647   & 2.569 &  1.511   & 0.718 & 0.653  &0.773  & 0.695    & 0.661   & \cellcolor{light-gray}0.594  & 0.671     \\
Lipop (4,200) & RMSE    & 0.683 & \cellcolor{light-gray}\textbf{0.585}   & \cellcolor{light-gray}0.626  & 1.016 & 1.207    & 0.947 &0.633  &0.665  & 0.669    & 0.680  & 0.664    &  \cellcolor{light-gray}0.630 \\
FreeSolv (639) & RMSE    &1.063  & \cellcolor{light-gray}\textbf{0.715}  &  \cellcolor{light-gray}1.014 & 2.275 &   2.205  & 1.504 & \cellcolor{light-gray}1.046  & 1.316 & 1.327    & 1.304  &  1.139    &  1.159 \\
QM7 (6,830) & MAE    & \cellcolor{light-gray}\textbf{42.814}  & \cellcolor{light-gray}52.726   &\cellcolor{light-gray}51.403   & 81.165 & 158.160    & 64.363 & 86.060  & 64.530 & 107.013    & 78.217  &   59.973   & 55.727   \\
QM8 (21,786) &MAE    &0.0364  &0.0126    & \cellcolor{light-gray}\textbf{0.0098}  & 0.0205 & 0.0295   & 0.0232 &0.0104   &0.0154  &0.0109     & 0.0187  & \cellcolor{light-gray}\textbf{0.0098}  &  \cellcolor{light-gray}0.0103     \\
\bottomrule
\end{tabular}}
\end{table*}
\begin{figure*}[t]
    \begin{center}
    \includegraphics[width=0.918\textwidth]{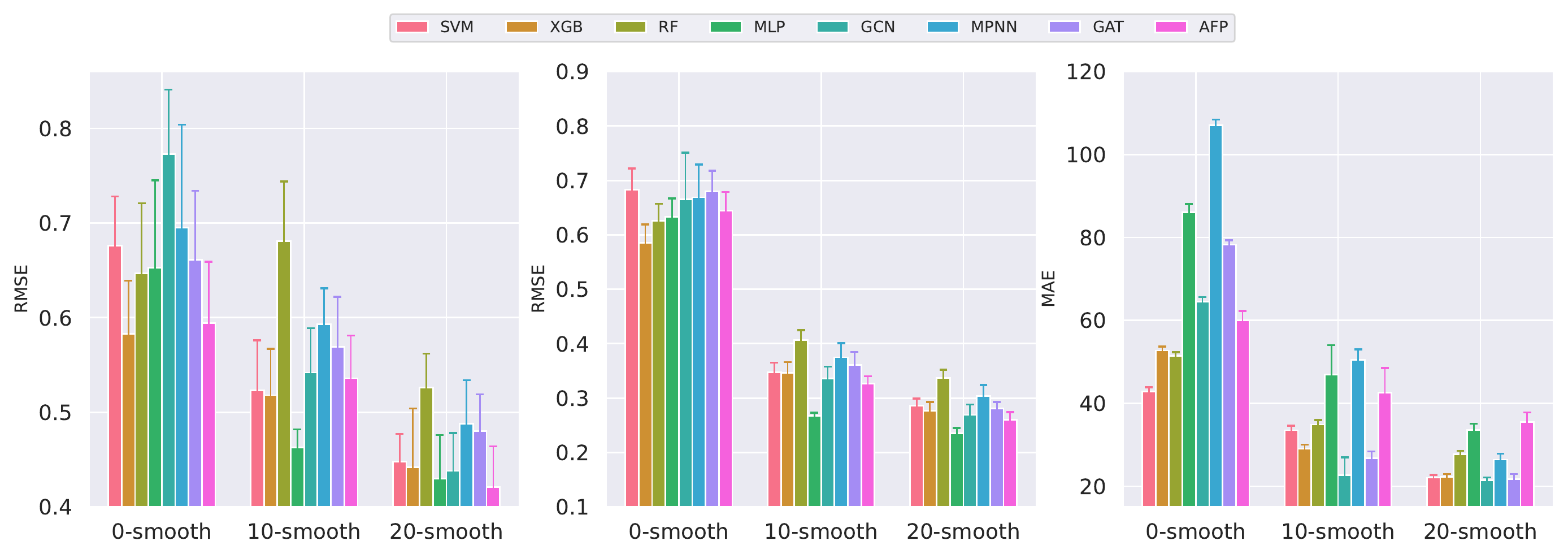}
    \end{center}
    \vspace{-1em}
    \caption{The performance of various models on the smoothed datasets. \textbf{Left}: ESOL (Regression); \textbf{Middle}: Lipop (Regression); \textbf{Right}: QM7 (Regression). We only smooth the regression datasets because the labels of classification datasets are not suitable for smoothing.}
   \label{fig_smooth}
   \vspace{-1em}
\end{figure*}
\subsection{Observations}
Table~\ref{tab:main_results} documents the benchmark results for various models and datasets, from which we can make the following \underline{\textit{\textbf{Observations}}}:

\underline{\textit{\textbf{Observation 1.}}} \textbf{Deep models underperform non-deep counterparts in most cases.}\\
As can be observed in Table~\ref{tab:main_results}, non-deep models rank as the top one on 10/14 datasets. On some datasets such as MUV, QM7, and BACE, three non-deep models can even beat any deep models.

\underline{\textit{\textbf{Observation 2.}}} \textbf{It is irregular data patterns, NOT solely the small size of molecular datasets to blame for the failure of deep models!}\\
Intuitively, many previous works~\cite{goh2017deep,yang2019analyzing} pointed out that the small size of molecular datasets could be a bottleneck for deep learning models. Here, we provide a second voice to such pre-dominant beliefs with empirical evidence. As shown in Table~\ref{tab:main_results}, all the non-deep models can outperform any deep ones on some larger-scale datasets (e.g., MUV and QM 7). However, in some small datasets (e.g., ClinTox and ESOL), some deep models can beat partial non-deep ones. Therefore, what matters is the irregular molecule data pattern, not solely the dataset size. We will provide an in-depth analysis to the unique molecule data pattern in Sec.~\ref{why}. 

\underline{\textit{\textbf{Observation 3.}}} \textbf{Tree models (\textbf{XGB} and \textbf{RF}) exhibit a particular advantage over other models.}\\
In the experiments shown in Table~\ref{tab:main_results}, we can see that the tree-based models consistently rank among the top three on each dataset. Additionally, tree models rank as the top one on 8/15 datasets. We will explore why tree models are well-suited for molecular fingerprints in Sec.~\ref{why}.
\vspace{-1em}
\section{Why above phenomena would occur?}
\label{why}
\begin{table*}[t]
\caption{RMSE\textsubscript{nc} and RMSE\textsubscript{c} are the prediction RMSE on non-cliff molecules and cliff molecules, respectively. $\Delta\mathcal{R} = $ (RMSE\textsubscript{c} - RMSE\textsubscript{nc}) / RMSE\textsubscript{nc} $\times 100\%$. The top-3 performances and the best performance are highlighted with grey background and $\textbf{bold}$.}
\label{tab:main_ACs}
\setlength{\tabcolsep}{1.68pt}
\centering
\fontsize{7.8pt}{\baselineskip}\selectfont
\resizebox{0.688\textwidth}{!}{
\begin{tabular}{@{}ccccccccccccc@{}}
\toprule
\textbf{\begin{tabular}[c]{@{}c@{}}Target name\\ (Response type)\end{tabular}}     & \textbf{Metric} & \textbf{SVM} & \textbf{XGB} & \textbf{RF} &\textbf{CNN} & \textbf{RNN} & \textbf{TRSF}  & \textbf{MLP} & \textbf{GCN} & \textbf{MPNN} & \textbf{GAT} & \textbf{AFP} 
\\ \midrule

\multirow{3}{*}{\begin{tabular}[c]{@{}c@{}}CB1\\ (Agonism EC\textsubscript{50})\end{tabular}}      & RMSE\textsubscript{nc}    &  \cellcolor{light-gray}0.652        &  \cellcolor{light-gray}0.623            &  \cellcolor{light-gray}\textbf{0.619}  & 0.934 & 0.712  & 0.785    & 0.707        & 0.932        & 0.938         & 0.960          & 0.909             \\
 & RMSE\textsubscript{c}     &  \cellcolor{light-gray}0.773        &  \cellcolor{light-gray}\textbf{0.767}            &  \cellcolor{light-gray}0.770 & 0.944 & 0.823  & 0.888     & 0.807        & 0.992        & 0.989         & 0.975       & 0.967            \\
&   $\Delta\mathcal{R}$           & 18.55$\%$    &  23.11$\%$        &  24.39$\%$ & 1.15$\%$  & 15.59$\%$  &13.12$\%$  & 14.1$\%$      & 6.37$\%$      & 5.47$\%$       & 1.55$\%$         & 6.35$\%$            \\ \midrule
                                                                     
\multirow{3}{*}{\begin{tabular}[c]{@{}c@{}}DAT\\ (Inhibition K\textsubscript{i})\end{tabular}}         & RMSE\textsubscript{nc}    &  \cellcolor{light-gray}0.589        &  \cellcolor{light-gray}0.579            & \cellcolor{light-gray} \textbf{0.577} & 0.871  & 0.692 & 0.801    & 0.664        & 0.927        & 0.820         & 0.995          & 0.865                \\
 & RMSE\textsubscript{c}     &  \cellcolor{light-gray}0.744        & \cellcolor{light-gray} \textbf{0.696}            &  \cellcolor{light-gray}0.730   & 0.894&0.783  & 0.934       & 0.792        & 1.003        & 0.921         & 1.042          & 0.995                \\ 
&   $\Delta\mathcal{R}$           & 26.30$\%$      & 20.18$\%$   & 26.64$\%$  &2.48$\%$              & 13.15$\%$  & 16.70$\%$   & 19.40$\%$       & 8.23$\%$        & 12.38$\%$        & 4.74$\%$          & 15.11$\%$               \\ \midrule
\multirow{3}{*}{\begin{tabular}[c]{@{}c@{}}PPAR$\alpha$\\ (Agonism EC\textsubscript{50})\end{tabular}} & RMSE\textsubscript{nc}    &  \cellcolor{light-gray}\textbf{0.535}        &  \cellcolor{light-gray}0.552    &  \cellcolor{light-gray}0.561 & 0.854 & 0.696  & 0.799         & 0.606        & 0.856        & 0.833         & 0.892          & 0.749                \\
& RMSE\textsubscript{c}     &  \cellcolor{light-gray}\textbf{0.671}        &  \cellcolor{light-gray}0.678            &  \cellcolor{light-gray}0.685 & 0.962  & 0.825 &  0.968      & 0.713        & 0.870        & 0.872         & 0.929          & 0.823                \\
  &  $\Delta\mathcal{R}$         &  25.42$\%$       & 22.83$\%$   & 22.10$\%$  &12.69$\%$          & 15.64$\%$   &  21.26$\%$     & 17.77$\%$       & 1.72$\%$        & 4.78$\%$         & 4.21$\%$          & 9.90$\%$                \\ \midrule
  
\multirow{3}{*}{\begin{tabular}[c]{@{}c@{}}DOR\\ (Inhibition K\textsubscript{i})\end{tabular}}      & RMSE\textsubscript{nc}   &  \cellcolor{light-gray}0.598         &  \cellcolor{light-gray}0.592   &  \cellcolor{light-gray}\textbf{0.591}          & 0.938 & 0.893  & 0.873       & 0.663        & 1.095        & 0.958         & 1.102          & 1.018                \\
& RMSE\textsubscript{c}         &  \cellcolor{light-gray}0.861    &  \cellcolor{light-gray}0.854           &  \cellcolor{light-gray}\textbf{0.836}   & 1.098 & 1.036 &  1.032   & 0.874        & 1.259        & 1.152         & 1.281          & 1.179                \\
 &  $\Delta\mathcal{R}$        & 43.98$\%$       & 44.14$\%$   & 41.46$\%$       & 17.06$\%$  & 16.01 $\%$  & 18.26$\%$      & 31.85$\%$       & 14.93$\%$       & 20.27$\%$         & 16.26$\%$         & 15.83$\%$               \\   
                                                        \bottomrule
\end{tabular}
}
\end{table*}
In this section, we attempt to understand which characteristics of molecular data lead to the failure of powerful deep models. Also, we aim to understand the inductive biases of tree models that make them well-suited for molecules, and how they differ from the inductive biases of deep models.

\underline{\textit{\textbf{Explanation 1.}}} \textbf{Unlike image data, molecular data patterns are non-smooth. Deep models struggle to learn non-smooth target functions that map molecules to properties.}\\
We design two experiments to verify the above explanation, i.e., increasing or decreasing the level of data smoothing in the molecular datasets. Firstly, we transform the molecular data by smoothing the labels based on similarities between molecules. Specifically, let $\mathcal{D}$ denote the molecular dataset and $(x_i, y_i)\in\mathcal{D}$ be $i$-th molecule and its label, we smooth the target function as follows,
\begin{equation}
    \widehat{y_i} = \frac{\sum_{x_{j}\in\mathcal{N}_{x_i}}s(x_{i}, x_{j})y_{j}}{\sum_{x_{j}\in\mathcal{N}_{x_i}}s(x_{i}, x_{j})},
\end{equation}
where $s(\cdot,\cdot)$ denotes the Tanimoto coefficient of the extended connectivity fingerprints (ECFP) between two molecules that can be considered as their structural similarity. $\mathcal{N}_{x_i}$ is the $k$-nearest neighbor set of $x_i$ (including $x_i$) picked from the whole dataset based on the structural similarities. $\widehat{y_i}$ denotes the label after smoothing. We smooth all the molecules in the dataset in this way and use the smoothed label $\widehat{y_i}$ to train the models. The results are shown in Figure~\ref{fig_smooth}, where `0-smooth' denotes the original datasets. `10-smooth' and `20-smooth' mean $k=10$ and $k=20$, respectively. As can be observed, the performance of deep models improves dramatically as the level of dataset smoothing increases, and many deep models including MLP, GCN, and AFP can even beat non-deep ones after smoothing. These phenomena indicate that deep models are more suitable for the smoothed datasets.

\begin{figure}[h!]
    \begin{center}
    \includegraphics[width=0.418\textwidth]{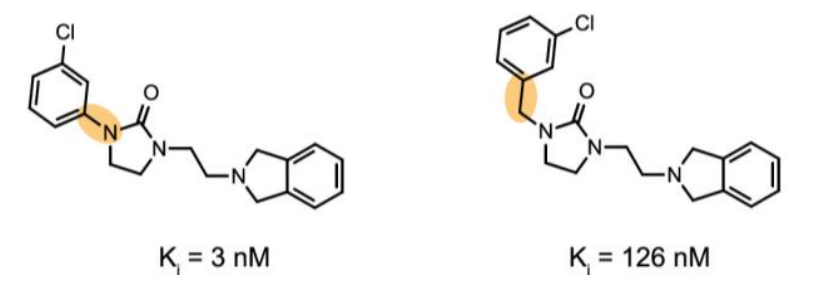}
    \end{center}
    \vspace{-1em}
    \caption{Examplary of Activity Cliffs (ACs) on the target named dopamine D3 receptor (D3R). K$_i$ means the bioactivity values. This figure is adapted from Derek van Tilborg's work~\cite{van2022exposing}.}
   \label{fig_AC}
\end{figure}
Secondly, we decrease the level of data smoothing using the concept of activity cliff~\cite{maggiora2006outliers,stumpfe2012exploring} from chemistry, which means a situation where small changes in the chemical structure of a drug lead to significant changes in its bioactivity. We provide an example activity cliff pairs in Figure~\ref{fig_AC}. Apparently, the target function of activity cliffs that map molecules to the activity values is less smooth than normal molecular datasets. We then evaluate the models on the activity cliff datasets~\cite{van2022exposing}. The test set contains molecules that are chemically similar to those in the training set but exhibit either a large difference in bioactivity (cliff molecules) or similar bioactivity (non-cliff molecules).
As shown in Table~\ref{tab:main_ACs}, the non-deep models consistently outperform deep ones on these activity cliff datasets. Furthermore, it is worth noting that deep models exhibit a similar level of performance on both non-cliff and cliff molecules, while non-deep models experience significant changes in performance when transitioning from non-cliff to cliff molecules. These phenomena indicate that deep models are less sensitive to subtle structural changes and struggle to learn non-smooth target functions compared with tree models, especially the activity cliff cases. Our explanation is consistent with the conclusions in deep learning theory~\cite{rahaman2019spectral}, i.e., deep models struggle to learn high-frequency components of the target functions. However, tree models can learn piece-wise target functions, and do not exhibit such  bias. Our explorations uncover several promising avenues to enhance deep models' performance on molecules: smoothing the target functions or improving deep models' ability to learn the non-smooth target functions.

\underline{\textit{\textbf{Explanation 2.}}} \textbf{Deep models mix different dimensions of molecular features, whereas tree models make decisions based on each dimension of the features separately.}
\\
Typically, features in molecular data carry meanings individually. Each dimension of molecular fingerprints often indicates whether a certain substructure is present in the molecule; each dimension of nodes/edges features in molecular graph data indicates a specific characteristic of the atoms/bonds (e.g., atom/bond type, atom degree).
To verify the above explanation, we mix the different dimensions of molecular features $x_i \in \mathbb{R}^{d}$ using an orthogonal transformation before feeding them into various models,
\begin{equation}
    \widehat{x_i} = \mathcal{Q}x_i, 
\end{equation}
where $\mathcal{Q} \in \mathbb{R}^{d\times d}$ is the orthogonal matrix and $\widehat{x_i}$ is the molecular feature after transformation. Kindly note that the meaning of $x_i$ depends on the input molecular descriptors in the experiments. Specifically, for SVM, XGB, RF, and MLP, $x_i$ denotes the molecular fingerprints; for GNN models, $x_i$ can denote the atom features and bond features in the molecular graphs, i.e., we apply orthogonal transformations to both the atom features and bond features. As can be observed in Figure~\ref{fig_rot}, the performance of tree models deteriorates dramatically and falls behind most deep models after the orthogonal transformation. It is because each dimension of $\widehat{x_i}$ is a convex combination of all the dimensions of $x_i$ according to the matrix-vector product rule. In other words, the molecular features after orthogonal transformation no longer carry meanings individually, accounting for the failure of tree models that make decisions based on each dimension of the features separately. The learning style of tree models is more suitable for molecular data because only a handful of features (e.g., certain substructures) are most indicative of molecular properties. On the other hand, the performance decreases of deep models are less significant, and most deep models can beat tree models after the transformations. We explain this observation as follows. Without the loss of generality, we assume that a linear layer of deep models can map the original molecular feature $x_i$ to the label $y_i$,
\begin{equation}
    y_i =  W^{\top} x_i + b, 
\end{equation}
where $W$ and $b$ denote the parameter matrix and the bias term of the linear layer, respectively. And then, we aim to learn a new linear layer mapping the transformed model feature $\widehat{x_i}$ to label $y_i$,
\begin{equation}
    y_i = \widehat{W}^{\top}\widehat{x_i}+b = \widehat{W}^{\top}\mathcal{Q}x_i+\hat{b},
\end{equation}
where $\widehat{W}$ and $\hat{b}$ denote the parameter matrix and the bias term of the new linear layer, respectively. Apparently, to achieve the same results as the original feature, we only have to learn $\widehat{W}$ so that $\widehat{W} = \mathcal{Q}W$ because
$\mathcal{Q}^{-1} = \mathcal{Q}^{\top}$ as an orthogonal matrix, and also $\hat{b} = b$. Therefore, applying the orthogonal transformation to molecular features barely impacts the performance of deep models. The empirical results in Figure~\ref{fig_rot} confirm this point although some performance changes are observable due to uncontrollable random factors.
This explanation inspires us not to mix the molecular features before feeding them into models.

\bibliography{example_paper}

\begin{thebibliography}{78}
\providecommand{\natexlab}[1]{#1}
\providecommand{\url}[1]{\texttt{#1}}
\expandafter\ifx\csname urlstyle\endcsname\relax
  \providecommand{\doi}[1]{doi: #1}\else
  \providecommand{\doi}{doi: \begingroup \urlstyle{rm}\Url}\fi

\bibitem[Atz et~al.(2021)Atz, Grisoni, et~al.]{atz2021geometric}
Atz, K., Grisoni, F., et~al.
\newblock {Geometric deep learning on molecular representations}.
\newblock \emph{{Nature Machine Intelligence}}, 2021.

\bibitem[Axen et~al.(2017)Axen, Huang, C{\'a}ceres, Gendelev, Roth, and
  Keiser]{axen2017simple}
Axen, S.~D., Huang, X.-P., C{\'a}ceres, E.~L., Gendelev, L., Roth, B.~L., and
  Keiser, M.~J.
\newblock A simple representation of three-dimensional molecular structure.
\newblock \emph{Journal of medicinal chemistry}, 60\penalty0 (17):\penalty0
  7393--7409, 2017.

\bibitem[Chen \& Guestrin(2016)Chen and Guestrin]{chen2016xgboost}
Chen, T. and Guestrin, C.
\newblock Xgboost: A scalable tree boosting system.
\newblock In \emph{Proceedings of the 22nd acm sigkdd international conference
  on knowledge discovery and data mining}, pp.\  785--794, 2016.

\bibitem[Cho et~al.(2014)Cho, Van~Merri{\"e}nboer, Gulcehre, Bahdanau,
  Bougares, Schwenk, and Bengio]{cho2014learning}
Cho, K., Van~Merri{\"e}nboer, B., Gulcehre, C., Bahdanau, D., Bougares, F.,
  Schwenk, H., and Bengio, Y.
\newblock Learning phrase representations using rnn encoder-decoder for
  statistical machine translation.
\newblock \emph{arXiv preprint arXiv:1406.1078}, 2014.

\bibitem[Coley et~al.(2017)Coley, Barzilay, et~al.]{coley2017convolutional}
Coley, C.~W., Barzilay, R., et~al.
\newblock {Convolutional embedding of attributed molecular graphs for physical
  property prediction}.
\newblock \emph{{Journal of chemical information and modeling}}, 2017.

\bibitem[Devlin et~al.(2019)Devlin, Chang, et~al.]{devlin2019bert}
Devlin, J., Chang, M., et~al.
\newblock {BERT: Pre-training of Deep Bidirectional Transformers for Language
  Understanding}.
\newblock In \emph{{NAACL}}, 2019.

\bibitem[Du et~al.(2022)Du, Zhang, et~al.]{du2022se}
Du, W., Zhang, H., et~al.
\newblock {SE(3) Equivariant Graph Neural Networks with Complete Local Frames}.
\newblock In \emph{{ICML}}, 2022.

\bibitem[Duvenaud et~al.(2015)Duvenaud, Maclaurin, Iparraguirre, Bombarell,
  Hirzel, Aspuru-Guzik, and Adams]{duvenaud2015convolutional}
Duvenaud, D.~K., Maclaurin, D., Iparraguirre, J., Bombarell, R., Hirzel, T.,
  Aspuru-Guzik, A., and Adams, R.~P.
\newblock Convolutional networks on graphs for learning molecular fingerprints.
\newblock \emph{Advances in neural information processing systems}, 28, 2015.

\bibitem[Gao et~al.(2022)Gao, Tan, Wu, and Li]{gao2022cosp}
Gao, Z., Tan, C., Wu, L., and Li, S.~Z.
\newblock Cosp: Co-supervised pretraining of pocket and ligand.
\newblock \emph{arXiv preprint arXiv:2206.12241}, 2022.

\bibitem[Gasteiger et~al.(2020)Gasteiger, Groß, and
  Günnemann]{Gasteiger2020Directional}
Gasteiger, J., Groß, J., and Günnemann, S.
\newblock Directional message passing for molecular graphs.
\newblock In \emph{International Conference on Learning Representations}, 2020.
\newblock URL \url{https://openreview.net/forum?id=B1eWbxStPH}.

\bibitem[Gasteiger et~al.(2021)Gasteiger, Becker, and
  G{\"u}nnemann]{gasteiger2021gemnet}
Gasteiger, J., Becker, F., and G{\"u}nnemann, S.
\newblock Gemnet: Universal directional graph neural networks for molecules.
\newblock \emph{Advances in Neural Information Processing Systems},
  34:\penalty0 6790--6802, 2021.

\bibitem[Gilmer et~al.(2017)Gilmer, Schoenholz, Riley, Vinyals, and
  Dahl]{gilmer2017neural}
Gilmer, J., Schoenholz, S.~S., Riley, P.~F., Vinyals, O., and Dahl, G.~E.
\newblock Neural message passing for quantum chemistry.
\newblock In \emph{Proceedings of the 34th International Conference on Machine
  Learning}, pp.\  1263--1272, 2017.

\bibitem[Goh et~al.(2017)Goh, Hodas, and Vishnu]{goh2017deep}
Goh, G.~B., Hodas, N.~O., and Vishnu, A.
\newblock Deep learning for computational chemistry.
\newblock \emph{Journal of computational chemistry}, 38\penalty0 (16):\penalty0
  1291--1307, 2017.

\bibitem[Gu et~al.(2018)Gu, Wang, Kuen, Ma, Shahroudy, Shuai, Liu, Wang, Wang,
  Cai, et~al.]{gu2018recent}
Gu, J., Wang, Z., Kuen, J., Ma, L., Shahroudy, A., Shuai, B., Liu, T., Wang,
  X., Wang, G., Cai, J., et~al.
\newblock Recent advances in convolutional neural networks.
\newblock \emph{Pattern recognition}, 77:\penalty0 354--377, 2018.

\bibitem[Hamilton et~al.(2017)Hamilton, Ying, and
  Leskovec]{hamilton2017inductive}
Hamilton, W., Ying, Z., and Leskovec, J.
\newblock Inductive representation learning on large graphs.
\newblock \emph{Advances in neural information processing systems}, 30, 2017.

\bibitem[Honda et~al.(2019)Honda, Shi, and Ueda]{honda2019smiles}
Honda, S., Shi, S., and Ueda, H.~R.
\newblock Smiles transformer: Pre-trained molecular fingerprint for low data
  drug discovery.
\newblock \emph{arXiv preprint arXiv:1911.04738}, 2019.

\bibitem[Hu et~al.(2022)Hu, Xia, Zheng, Tan, Huang, Xu, and Li]{hu2022protein}
Hu, B., Xia, J., Zheng, J., Tan, C., Huang, Y., Xu, Y., and Li, S.~Z.
\newblock Protein language models and structure prediction: Connection and
  progression.
\newblock \emph{arXiv preprint arXiv:2211.16742}, 2022.

\bibitem[Hu et~al.(2020)Hu, Liu, and others.]{hu2020strategies}
Hu, W., Liu, B., and others.
\newblock Strategies for pre-training graph neural networks.
\newblock \emph{ICLR}, 2020.

\bibitem[Janela \& Bajorath(2022)Janela and Bajorath]{janela2022simple}
Janela, T. and Bajorath, J.
\newblock Simple nearest-neighbour analysis meets the accuracy of compound
  potency predictions using complex machine learning models.
\newblock \emph{Nature Machine Intelligence}, pp.\  1--10, 2022.

\bibitem[Jiang et~al.(2021)Jiang, Wu, Hsieh, Chen, Liao, Wang, Shen, Cao, Wu,
  and Hou]{jiang2021could}
Jiang, D., Wu, Z., Hsieh, C.-Y., Chen, G., Liao, B., Wang, Z., Shen, C., Cao,
  D., Wu, J., and Hou, T.
\newblock Could graph neural networks learn better molecular representation for
  drug discovery? a comparison study of descriptor-based and graph-based
  models.
\newblock \emph{Journal of cheminformatics}, 13\penalty0 (1):\penalty0 1--23,
  2021.

\bibitem[Kearnes et~al.(2016)Kearnes, McCloskey, et~al.]{kearnes2016molecular}
Kearnes, S., McCloskey, K., et~al.
\newblock {Molecular Graph Convolutions: Moving Beyond Fingerprints}.
\newblock \emph{{J. Comput. Aided Mol. Des.}}, 2016.

\bibitem[Kimber et~al.(2021)Kimber, Gagnebin, and Volkamer]{kimber2021maxsmi}
Kimber, T.~B., Gagnebin, M., and Volkamer, A.
\newblock Maxsmi: maximizing molecular property prediction performance with
  confidence estimation using smiles augmentation and deep learning.
\newblock \emph{Artificial Intelligence in the Life Sciences}, 1:\penalty0
  100014, 2021.

\bibitem[Kipf \& Welling(2017)Kipf and Welling]{kipf2017semi-supervised}
Kipf, N.~T. and Welling, M.
\newblock {Semi-Supervised Classification with Graph Convolutional Networks}.
\newblock In \emph{{ICLR}}, 2017.

\bibitem[Krenn et~al.(2020)Krenn, H{\"a}se, Nigam, Friederich, and
  Aspuru-Guzik]{krenn2020self}
Krenn, M., H{\"a}se, F., Nigam, A., Friederich, P., and Aspuru-Guzik, A.
\newblock Self-referencing embedded strings (selfies): A 100\% robust molecular
  string representation.
\newblock \emph{Machine Learning: Science and Technology}, 1\penalty0
  (4):\penalty0 045024, 2020.

\bibitem[Landrum(2013)]{landrum2013rdkit}
Landrum, G.
\newblock {RDKit: A software suite for cheminformatics, computational
  chemistry, and predictive modeling}, 2013.

\bibitem[Li et~al.(2021)Li, Wang, and others.]{li2021an}
Li, P., Wang, J., and others.
\newblock An effective self-supervised framework for learning expressive
  molecular global representations to drug discovery.
\newblock \emph{BIB}, 2021.

\bibitem[Liu et~al.(2021)Liu, Wang, et~al.]{liu2021spherical}
Liu, Y., Wang, L., et~al.
\newblock {Spherical message passing for 3d molecular graphs}.
\newblock In \emph{{Iclr}}, 2021.

\bibitem[Liu et~al.(2022)Liu, Wang, Liu, Lin, Zhang, Oztekin, and
  Ji]{liu2022spherical}
Liu, Y., Wang, L., Liu, M., Lin, Y., Zhang, X., Oztekin, B., and Ji, S.
\newblock Spherical message passing for 3d molecular graphs.
\newblock In \emph{International Conference on Learning Representations
  (ICLR)}, 2022.

\bibitem[Liu et~al.(2023)Liu, Liang, Xia, Zhou, Yang, Liu, and Li]{liu2023dink}
Liu, Y., Liang, K., Xia, J., Zhou, S., Yang, X., Liu, X., and Li, S.~Z.
\newblock Dink-net: Neural clustering on large graphs.
\newblock \emph{arXiv preprint arXiv:2305.18405}, 2023.

\bibitem[Lu et~al.(2019)Lu, Liu, Wang, Huang, Lin, and He]{lu2019molecular}
Lu, C., Liu, Q., Wang, C., Huang, Z., Lin, P., and He, L.
\newblock Molecular property prediction: A multilevel quantum interactions
  modeling perspective.
\newblock In \emph{Proceedings of the AAAI Conference on Artificial
  Intelligence}, volume~33, pp.\  1052--1060, 2019.

\bibitem[Maggiora(2006)]{maggiora2006outliers}
Maggiora, G.~M.
\newblock On outliers and activity cliffs why qsar often disappoints, 2006.

\bibitem[Mayr et~al.(2018)Mayr, Klambauer, Unterthiner, Steijaert, Wegner,
  Ceulemans, Clevert, and Hochreiter]{mayr2018large}
Mayr, A., Klambauer, G., Unterthiner, T., Steijaert, M., Wegner, J.~K.,
  Ceulemans, H., Clevert, D.-A., and Hochreiter, S.
\newblock Large-scale comparison of machine learning methods for drug target
  prediction on chembl.
\newblock \emph{Chemical science}, 9\penalty0 (24):\penalty0 5441--5451, 2018.

\bibitem[Medsker \& Jain(1999)Medsker and Jain]{medsker1999recurrent}
Medsker, L. and Jain, L.~C.
\newblock \emph{Recurrent neural networks: design and applications}.
\newblock CRC press, 1999.

\bibitem[Min et~al.(2022)Min, Chen, Bian, Xu, Zhao, Huang, Zhao, Huang,
  Ananiadou, and Rong]{min2022transformer}
Min, E., Chen, R., Bian, Y., Xu, T., Zhao, K., Huang, W., Zhao, P., Huang, J.,
  Ananiadou, S., and Rong, Y.
\newblock Transformer for graphs: An overview from architecture perspective.
\newblock \emph{arXiv preprint arXiv:2202.08455}, 2022.

\bibitem[Mulder et~al.(2015)Mulder, Bethard, and Moens]{Mulder2015ASO}
Mulder, W.~D., Bethard, S., and Moens, M.-F.
\newblock A survey on the application of recurrent neural networks to
  statistical language modeling.
\newblock \emph{Comput. Speech Lang.}, 30:\penalty0 61--98, 2015.

\bibitem[Ozaki et~al.(2020)Ozaki, Tanigaki, Watanabe, and
  Onishi]{Ozaki2020MultiobjectiveTP}
Ozaki, Y., Tanigaki, Y., Watanabe, S., and Onishi, M.
\newblock Multiobjective tree-structured parzen estimator for computationally
  expensive optimization problems.
\newblock \emph{Proceedings of the 2020 Genetic and Evolutionary Computation
  Conference}, 2020.

\bibitem[Ozturk et~al.(2018)Ozturk, Ozgur, and Ozkirimli]{ozturk2018deepdta}
Ozturk, H., Ozgur, A., and Ozkirimli, E.
\newblock Deepdta: deep drug--target binding affinity prediction.
\newblock \emph{Bioinformatics}, 34\penalty0 (17):\penalty0 i821--i829, 2018.

\bibitem[Pattanaik \& Coley(2020)Pattanaik and Coley]{pattanaik2020molecular}
Pattanaik, L. and Coley, C.~W.
\newblock Molecular representation: going long on fingerprints.
\newblock \emph{Chem}, 6\penalty0 (6):\penalty0 1204--1207, 2020.

\bibitem[Rahaman et~al.(2019)Rahaman, Baratin, Arpit, Draxler, Lin, Hamprecht,
  Bengio, and Courville]{rahaman2019spectral}
Rahaman, N., Baratin, A., Arpit, D., Draxler, F., Lin, M., Hamprecht, F.,
  Bengio, Y., and Courville, A.
\newblock On the spectral bias of neural networks.
\newblock In \emph{International Conference on Machine Learning}, pp.\
  5301--5310. PMLR, 2019.

\bibitem[Rogers \& Hahn(2010)Rogers and Hahn]{rogers2010extended-connectivity}
Rogers, D. and Hahn, M.
\newblock Extended-connectivity fingerprints.
\newblock \emph{J chem inf}, 2010.

\bibitem[Rong et~al.(2020)Rong, Bian, Xu, Xie, Wei, Huang, and
  Huang]{rong2020self}
Rong, Y., Bian, Y., Xu, T., Xie, W., Wei, Y., Huang, W., and Huang, J.
\newblock Self-supervised graph transformer on large-scale molecular data.
\newblock \emph{Advances in Neural Information Processing Systems},
  33:\penalty0 12559--12571, 2020.

\bibitem[Ross et~al.(2022)Ross, Belgodere, et~al.]{ross2022molformer}
Ross, J., Belgodere, B., et~al.
\newblock {Molformer: Large Scale Chemical Language Representations Capture
  Molecular Structure and Properties}.
\newblock \emph{{Nat. Mach. Intell.}}, 2022.

\bibitem[Satorras et~al.(2021)Satorras, Hoogeboom, et~al.]{satorras2021n}
Satorras, V.~G., Hoogeboom, E., et~al.
\newblock {E(n) Equivariant Graph Neural Networks}.
\newblock In \emph{{ICML}}, 2021.

\bibitem[Schuett et~al.(2017)Schuett, Kindermans, and
  others.]{schuett2017schnet}
Schuett, T.~K., Kindermans, and others.
\newblock Schnet: A continuous-filter convolutional neural network for modeling
  quantum interactions.
\newblock \emph{NIPS}, 2017.

\bibitem[Song et~al.(2020)Song, Zheng, Niu, Fu, Lu, and
  Yang]{song2020communicative}
Song, Y., Zheng, S., Niu, Z., Fu, Z.-H., Lu, Y., and Yang, Y.
\newblock Communicative representation learning on attributed molecular graphs.
\newblock In \emph{IJCAI}, volume 2020, pp.\  2831--2838, 2020.

\bibitem[Stumpfe \& Bajorath(2012)Stumpfe and Bajorath]{stumpfe2012exploring}
Stumpfe, D. and Bajorath, J.
\newblock Exploring activity cliffs in medicinal chemistry: miniperspective.
\newblock \emph{Journal of medicinal chemistry}, 55\penalty0 (7):\penalty0
  2932--2942, 2012.

\bibitem[Svetnik et~al.(2003)Svetnik, Liaw, Tong, Culberson, Sheridan, and
  Feuston]{svetnik2003random}
Svetnik, V., Liaw, A., Tong, C., Culberson, J.~C., Sheridan, R.~P., and
  Feuston, B.~P.
\newblock Random forest: a classification and regression tool for compound
  classification and qsar modeling.
\newblock \emph{Journal of chemical information and computer sciences},
  43\penalty0 (6):\penalty0 1947--1958, 2003.

\bibitem[Tan et~al.(2021)Tan, Xia, Wu, and Li]{tan2021co}
Tan, C., Xia, J., Wu, L., and Li, S.~Z.
\newblock Co-learning: Learning from noisy labels with self-supervision.
\newblock In \emph{Proceedings of the 29th ACM International Conference on
  Multimedia}, pp.\  1405--1413, 2021.

\bibitem[Tan et~al.(2023)Tan, Gao, Xia, Hu, and Li]{tan2023global}
Tan, C., Gao, Z., Xia, J., Hu, B., and Li, S.~Z.
\newblock Global-context aware generative protein design.
\newblock In \emph{ICASSP 2023-2023 IEEE International Conference on Acoustics,
  Speech and Signal Processing (ICASSP)}, pp.\  1--5. IEEE, 2023.

\bibitem[Tian et~al.(2022)Tian, Ketkar, and Tao]{tian2022admetboost}
Tian, H., Ketkar, R., and Tao, P.
\newblock Admetboost: a web server for accurate admet prediction.
\newblock \emph{Journal of Molecular Modeling}, 28\penalty0 (12):\penalty0
  1--6, 2022.

\bibitem[Todeschini \& Consonni(2010)Todeschini and
  Consonni]{todeschini2010molecular}
Todeschini, R. and Consonni, V.
\newblock Molecular descriptors.
\newblock \emph{Recent Advances in QSAR Studies}, pp.\  29--102, 2010.

\bibitem[Valsecchi et~al.(2022)Valsecchi, Collarile, Grisoni, Todeschini,
  Ballabio, and Consonni]{valsecchi2022predicting}
Valsecchi, C., Collarile, M., Grisoni, F., Todeschini, R., Ballabio, D., and
  Consonni, V.
\newblock Predicting molecular activity on nuclear receptors by multitask
  neural networks.
\newblock \emph{Journal of Chemometrics}, 36\penalty0 (2):\penalty0 e3325,
  2022.

\bibitem[van Tilborg et~al.(2022)van Tilborg, Alenicheva, and
  Grisoni]{van2022exposing}
van Tilborg, D., Alenicheva, A., and Grisoni, F.
\newblock Exposing the limitations of molecular machine learning with activity
  cliffs.
\newblock \emph{Journal of Chemical Information and Modeling}, 62\penalty0
  (23):\penalty0 5938--5951, 2022.

\bibitem[Vaswani et~al.(2017)Vaswani, Shazeer, Parmar, Uszkoreit, Jones, Gomez,
  Kaiser, and Polosukhin]{vaswani2017attention}
Vaswani, A., Shazeer, N., Parmar, N., Uszkoreit, J., Jones, L., Gomez, A.~N.,
  Kaiser, {\L}., and Polosukhin, I.
\newblock Attention is all you need.
\newblock \emph{Advances in neural information processing systems}, 30, 2017.

\bibitem[Velickovic et~al.(2018)Velickovic, Cucurull,
  et~al.]{velickovic2018graph}
Velickovic, P., Cucurull, G., et~al.
\newblock {Graph Attention Networks}.
\newblock In \emph{{ICLR}}, 2018.

\bibitem[Wang et~al.(2019)Wang, Guo, and others.]{wang2019smiles-bert}
Wang, S., Guo, Y., and others.
\newblock Smiles-bert - large scale unsupervised pre-training for molecular
  property prediction.
\newblock \emph{BCB}, 2019.

\bibitem[Wang et~al.(2017)Wang, Bryant, et~al.]{wang2017pubchem}
Wang, Y., Bryant, S.~H., et~al.
\newblock {Pubchem Bioassay: 2017 Update}.
\newblock \emph{{Nucleic Acids Res.}}, 2017.

\bibitem[Weininger et~al.(1989)Weininger, Weininger, and
  Weininger]{weininger1989smiles.}
Weininger, D., Weininger, A., and Weininger, L.~J.
\newblock Smiles. 2. algorithm for generation of unique smiles notation.
\newblock \emph{JOURNAL OF CHEMICAL INFORMATION AND COMPUTER SCIENCES}, 1989.

\bibitem[Wu et~al.(2018)Wu, Ramsundar, Feinberg, Gomes, Geniesse, Pappu,
  Leswing, and Pande]{wu2018moleculenet}
Wu, Z., Ramsundar, B., Feinberg, E.~N., Gomes, J., Geniesse, C., Pappu, A.~S.,
  Leswing, K., and Pande, V.
\newblock Moleculenet: a benchmark for molecular machine learning.
\newblock \emph{Chemical science}, 9\penalty0 (2):\penalty0 513--530, 2018.

\bibitem[Xia et~al.(2021)Xia, Lin, et~al.]{xia2021towards}
Xia, J., Lin, H., et~al.
\newblock {Towards Robust Graph Neural Networks against Label Noise}, 2021.
\newblock URL \url{https://openreview.net/forum?id=H38f_9b90BO}.

\bibitem[Xia et~al.(2022{\natexlab{a}})Xia, Tan, Wu, Xu, and Li]{xia2022ot}
Xia, J., Tan, C., Wu, L., Xu, Y., and Li, S.~Z.
\newblock Ot cleaner: Label correction as optimal transport.
\newblock In \emph{ICASSP 2022-2022 IEEE International Conference on Acoustics,
  Speech and Signal Processing (ICASSP)}, pp.\  3953--3957. IEEE,
  2022{\natexlab{a}}.

\bibitem[Xia et~al.(2022{\natexlab{b}})Xia, Wu, , Chen, Hu, and
  Li]{xia2022simgrace}
Xia, J., Wu, L., , Chen, J., Hu, B., and Li, S.~Z.
\newblock {SimGRACE: A Simple Framework for Graph Contrastive Learning without
  Data Augmentation}.
\newblock In \emph{Proceedings of The Web Conference 2022}. Association for
  Computing Machinery, 2022{\natexlab{b}}.

\bibitem[Xia et~al.(2022{\natexlab{c}})Xia, Wu, et~al.]{xia2022progcl}
Xia, J., Wu, L., et~al.
\newblock {ProGCL: Rethinking Hard Negative Mining in Graph Contrastive
  Learning}.
\newblock In \emph{{ICML}}, 2022{\natexlab{c}}.

\bibitem[Xia et~al.(2022{\natexlab{d}})Xia, Zheng, Tan, Wang, and
  Li]{xia2022towards}
Xia, J., Zheng, J., Tan, C., Wang, G., and Li, S.~Z.
\newblock Towards effective and generalizable fine-tuning for pre-trained
  molecular graph models.
\newblock \emph{bioRxiv}, 2022{\natexlab{d}}.

\bibitem[Xia et~al.(2022{\natexlab{e}})Xia, Zhu, Du, and Li]{xia2022survey}
Xia, J., Zhu, Y., Du, Y., and Li, S.~Z.
\newblock A survey of pretraining on graphs: Taxonomy, methods, and
  applications.
\newblock \emph{arXiv preprint arXiv:2202.07893}, 2022{\natexlab{e}}.

\bibitem[Xia et~al.(2023{\natexlab{a}})Xia, Zhao, et~al.]{xia2023mole-bert}
Xia, J., Zhao, C., et~al.
\newblock {Mole-BERT: Rethinking Pre-training Graph Neural Networks for
  Molecules}.
\newblock In \emph{{ICLR}}, 2023{\natexlab{a}}.

\bibitem[Xia et~al.(2023{\natexlab{b}})Xia, Zhu, Du, Liu, and
  Li]{xia2023systematic}
Xia, J., Zhu, Y., Du, Y., Liu, Y., and Li, S.~Z.
\newblock A systematic survey of chemical pre-trained models.
\newblock \emph{IJCAI}, 2023{\natexlab{b}}.

\bibitem[Xiong et~al.(2019)Xiong, Wang, Liu, Zhong, Wan, Li, Li, Luo, Chen,
  Jiang, et~al.]{xiong2019pushing}
Xiong, Z., Wang, D., Liu, X., Zhong, F., Wan, X., Li, X., Li, Z., Luo, X.,
  Chen, K., Jiang, H., et~al.
\newblock Pushing the boundaries of molecular representation for drug discovery
  with the graph attention mechanism.
\newblock \emph{Journal of medicinal chemistry}, 63\penalty0 (16):\penalty0
  8749--8760, 2019.

\bibitem[Xiong et~al.(2020)Xiong, Wang, and others.]{xiong2020pushing}
Xiong, Z., Wang, D., and others.
\newblock Pushing the boundaries of molecular representation for drug discovery
  with graph attention mechanism.
\newblock \emph{J Med Chem}, 2020.

\bibitem[Yang et~al.(2019)Yang, Swanson, and others.]{yang2019analyzing}
Yang, K., Swanson, K., and others.
\newblock Analyzing learned molecular representations for property prediction.
\newblock \emph{J CHEM INF MODEL}, 2019.

\bibitem[Yap(2011)]{yap2011padel}
Yap, C.~W.
\newblock Padel-descriptor: An open source software to calculate molecular
  descriptors and fingerprints.
\newblock \emph{Journal of computational chemistry}, 32\penalty0 (7):\penalty0
  1466--1474, 2011.

\bibitem[Ying et~al.(2021)Ying, Cai, et~al.]{ying2021do}
Ying, C., Cai, T., et~al.
\newblock {Do Transformers Really Perform Badly for Graph Representation?}
\newblock In \emph{{NeurIPS}}, 2021.

\bibitem[You et~al.(2020)You, Chen, and others.]{you2020graph}
You, Y., Chen, T., and others.
\newblock Graph contrastive learning with augmentations.
\newblock In \emph{NeurIPS}, 2020.

\bibitem[Yue et~al.(2022)Yue, Jun, Sihang, Siwei, Xifeng, Xihong, Ke, Wenxuan,
  Wang, et~al.]{yue2022survey}
Yue, L., Jun, X., Sihang, Z., Siwei, W., Xifeng, G., Xihong, Y., Ke, L.,
  Wenxuan, T., Wang, L.~X., et~al.
\newblock A survey of deep graph clustering: Taxonomy, challenge, and
  application.
\newblock \emph{arXiv preprint arXiv:2211.12875}, 2022.

\bibitem[Y{\"u}ksel et~al.(2023)Y{\"u}ksel, Ulusoy, {\"U}nl{\"u}, Deniz, and
  Do{\u{g}}an]{yuksel2023selformer}
Y{\"u}ksel, A., Ulusoy, E., {\"U}nl{\"u}, A., Deniz, G., and Do{\u{g}}an, T.
\newblock Selformer: Molecular representation learning via selfies language
  models.
\newblock \emph{arXiv preprint arXiv:2304.04662}, 2023.

\bibitem[Zernov et~al.(2003)Zernov, Balakin, Ivaschenko, Savchuk, and
  Pletnev]{zernov2003drug}
Zernov, V.~V., Balakin, K.~V., Ivaschenko, A.~A., Savchuk, N.~P., and Pletnev,
  I.~V.
\newblock Drug discovery using support vector machines. the case studies of
  drug-likeness, agrochemical-likeness, and enzyme inhibition predictions.
\newblock \emph{Journal of chemical information and computer sciences},
  43\penalty0 (6):\penalty0 2048--2056, 2003.

\bibitem[Zheng et~al.(2022)Zheng, Wang, Wang, Xia, Huang, Zhao, Zhang, and
  Li]{zheng2022using}
Zheng, J., Wang, Y., Wang, G., Xia, J., Huang, Y., Zhao, G., Zhang, Y., and Li,
  S.~Z.
\newblock Using context-to-vector with graph retrofitting to improve word
  embeddings.
\newblock \emph{arXiv preprint arXiv:2210.16848}, 2022.

\bibitem[Zheng et~al.(2019)Zheng, Yan, Yang, and Xu]{zheng2019identifying}
Zheng, S., Yan, X., Yang, Y., and Xu, J.
\newblock Identifying structure--property relationships through smiles syntax
  analysis with self-attention mechanism.
\newblock \emph{Journal of chemical information and modeling}, 59\penalty0
  (2):\penalty0 914--923, 2019.

\end{thebibliography}
\bibliographystyle{icml2023}

\newpage
\appendix
\onecolumn
\section{The performance of various models on the orthogonally transformed dataset}
\begin{figure*}[h]
    \begin{center}
    \includegraphics[width=0.918\textwidth]{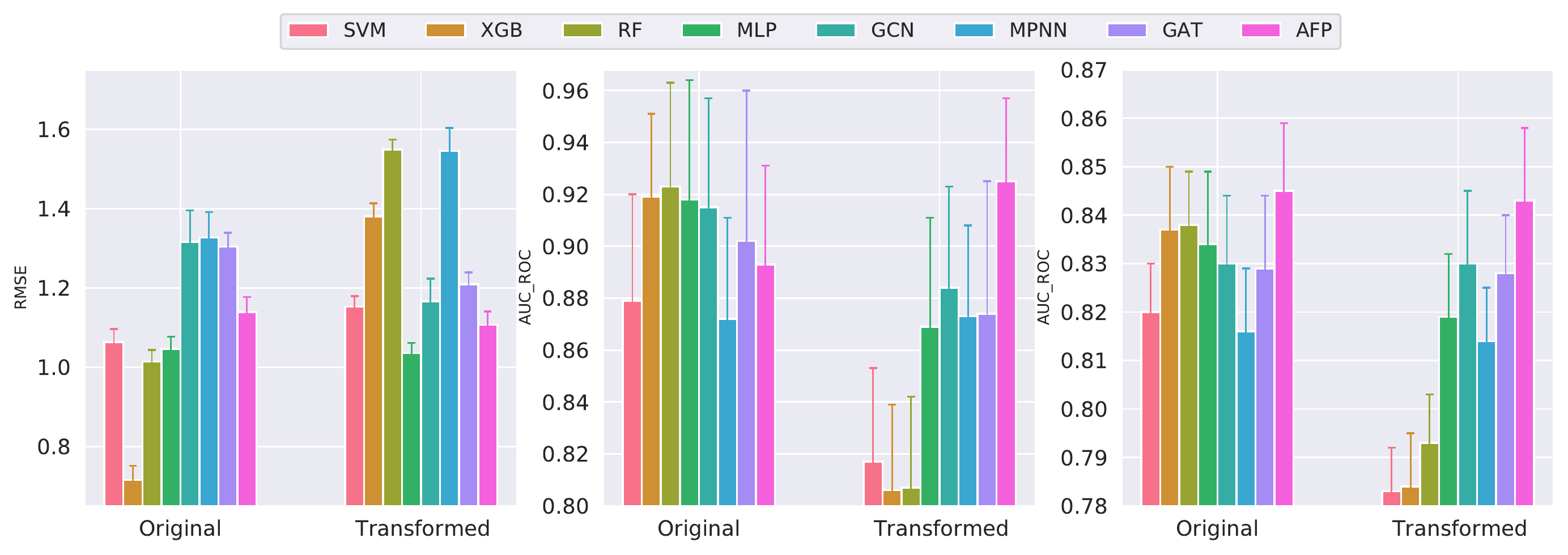}
    \end{center}
    \caption{The performance of various models on the orthogonally transformed datasets. \textbf{Left}: FreeSolv (Regression); \textbf{Middle}: ClinTox (Classification); \textbf{Right}: Tox21 (Classification). Kindly note that we did not evaluate CNN, RNN, and TRSF on the transformed datasets because we cannot apply the orthogonal transformations to the input SMILES strings.}
   \label{fig_rot}
\end{figure*}
\section{Experimental Setups}
\label{setup}
\textbf{Fingerprints $\longmapsto$ SVM, XGB, RF, and MLP.}\quad Following the common practice~\cite{tian2022admetboost,pattanaik2020molecular}, we feed the concatenation of various molecular fingerprints including 881 PubChem fingerprints (PubchemFP), 307 substructure fingerprints (SubFP), and 206 MOE 1-D and 2-D descriptors~\cite{yap2011padel} to SVM, XGB, RF, and MLP models to comprehensively represent molecular structures, with some pre-processing procedures to remove features (1) with missing values; (2) with extremely low variance (variance $\textless$ 0.05); (3) have a high correlation (Pearson correlation coefficient $\textgreater$ 0.95) with another feature. The retained features are normalized to the mean value of 0 and variance of 1. Additionally, considering that traditional machine models (SVM, RF, XGB) cannot be directly applied in the multi-task molecular datasets, we split the multi-task dataset into multiple single-task datasets and use each of them to train the models. Finally, we report the average performance of these single tasks.

\textbf{SMILES strings $\longmapsto$ CNN, RNN, and TRSF.}
We adopt the 1D CNNs from a recent study~\cite{kimber2021maxsmi}, which include a single 1D convolutional layer with a step size equal to 1,  followed by a fully connected layer. As for the RNN, we use a 3-layer bidirectional gated recurrent units (GRUs)~\cite{cho2014learning} with 256 hidden vector dimensions. Additionally, we use the pre-trained SMILES transformer~\cite{honda2019smiles} with 4 basic blocks and each block has 4-head attentions with 256 embedding dimensions and 2 linear layers. The SMILES are split into symbols (e.g., `Br', `C', `=', `(',`2') and then fed into the transformer together with the positional encoding~\cite{vaswani2017attention}. 

\textbf{2D Graphs $\longmapsto$ GCN, MPNN, GAT, and AFP.} 
As in previous studies~\cite{xiong2019pushing}, we exhaustively utilized all readily available atom/bond features in our 2D graph-based descriptors. Specifically, we have incorporated 9 atom features, including atom symbol, degree, and formal charge, using a one-hot encoding scheme. In addition, we included 4 bond features, such as type, conjugation, ring, and stereo. The resulting encoded graphs were then fed into GCN, MPNN, GAT, and AFP models. Further details on the graph descriptors used in our experiments can be found in~\cite{xiong2019pushing}.

\textbf{3D Graphs $\longmapsto$ SPN.} We employ the recently proposed SphereNet~\cite{liu2022spherical} for molecules with 3D geometry. Specifically, for quantum mechanics datasets (QM7 and QM8) that contain 3D atomic coordinates calculated with ab initio Density Functional Theory (DFT), we feed them into SphereNet directly. For other datasets without labeled conformations, we used RDKit~\cite{landrum2013rdkit}-generated conformations to satisfy the request of SphereNet.

\textbf{Datasets splits, evaluation protocols and metrics, hyper-parameters tuning.} Firstly, we randomly split the training, validation, and test sets at a ratio of 8:1:1. And then, we tune the hyper-parameters based on the performance of the validation set. Specifically, we select the optimal hyper-parameters set using the Tree of Parzen Estimators (TPE) algorithm~\cite{Ozaki2020MultiobjectiveTP} in 50 evaluations. Due to the heavy computational overhead, GNNs-based models on the HIV and MUV datasets are in 30 evaluations; all the models on the QM7 and QM8 are in 10 evaluations. And then, we conduct 50 independent runs with different random seeds for dataset splitting to obtain more reliable results, using the optimal hyper-parameters determined before. Similarly, GNNs-based models on the HIV and MUV datasets are in 30 evaluations; all the models on the QM7 and QM8 are in 10 evaluations. Following MoleculeNet benchmark~\cite{wu2018moleculenet}, we evaluate the classification tasks using the area under the receiver operating characteristic curve (AUC-ROC), except the area under the precision curve (AUC-PRC) on MUV dataset due to its extreme biased data distribution. The performance on the regression task are reported using root mean square error (RMSE) or mean absolute error (MAE). kindly note that we report the average performance across multi-tasks on some datasets because they contain more than one task. Additionally, to avoid the overfitting issue, all the deep models are trained with an early stopping scheme if no validation performance improvement is observed in successive 50 epochs. We set the maximal epoch as 300 and the batch-size as 128. 

\section{Related Work}
In this section, we elaborate on various molecular descriptors and their respective learning models.
\subsection{Fingerprints-based Molecular Descriptors}
\label{fp_related}
Molecular fingerprints (FPs) serve as one of the most important descriptors for molecules. Typical examples include Extended-Connectivity Fingerprints (ECFP)~\cite{rogers2010extended-connectivity} and PubChemFP~\cite{wang2017pubchem}. These fingerprints encode the neighboring environments of heavy atoms in a molecule into a fixed bit string with a hash function, where each bit indicates whether a certain substructure is present in the molecule. Traditional models (e.g., tree or SVM-based models) and MLPs can take these fingerprints as `raw' input. However, the high-dimensional and sparse nature of FPs introduces additional efforts for feature selection when they are fed into certain models. Additionally, it is difficult to interpret the relationship between properties and structures because the hash functions are non-invertible.
\subsection{Linear Notation-based Molecular Descriptors}
Another option for molecules is linear notations, among which SMILES~\cite{weininger1989smiles.} is the most frequently-used one owing to its versatility and interpretability. In SMILES, each atom is represented as a respective ASCII symbol; Chemical bonds, branching, and stereochemistry are denoted by specific symbols. However, a significant fraction of SMILES strings does not correspond to chemically valid molecules. As a remedy, a new language named SELF-referencIng Embedded Strings (SELFIES) for molecules was introduced in 2020~\cite{krenn2020self}. Every SELFIES string corresponds to a valid molecule, and
SELFIES can represent every molecule. Naturally, RNNs, 1D CNN, and Transformers are powerful deep models for processing such sequences~\cite{wang2019smiles-bert,zheng2019identifying,honda2019smiles,ross2022molformer,yuksel2023selformer}. However, the poor scalability of the sequential notations and the loss of spatial information limit the performances of these approaches.
\subsection{2D and 3D Graph-based Molecular Descriptors}
Molecules can be represented with graphs naturally, with nodes as atoms and edges as chemical bonds. Initially, \cite{duvenaud2015convolutional} first adopted convolutional layers to encode molecular graphs to neural fingerprints. Following this work, \cite{coley2017convolutional} employs the atom-based message-passing scheme to learn expressive molecular graph representations. To complement the atom's information, \cite{kearnes2016molecular} utilized both the atom's and bonds' attributes, and MPNN~\cite{gilmer2017neural} generalized it to a unified framework. Also, multiple variants of the MPNN framework are developed to avoid unnecessary loops (DMPNN~\cite{yang2019analyzing}), to strengthen the message interactions between nodes and edges (CMPNN~\cite{song2020communicative}), to capture the complex inherent quantum interactions of molecules (MGCN~\cite{lu2019molecular}),  or take the longer-range dependencies (Attentive FP~\cite{xiong2019pushing}). More recently, some hybrid architectures~\cite{rong2020self,ying2021do,min2022transformer} of GNNs and transformers are emerging to capture the topological structures of molecular graphs. Additionally, given that the available labels for molecules are often expensive or incorrect~\cite{xia2021towards,tan2021co,xia2022ot}, the emerging self-supervised pre-training strategies~\cite{you2020graph,xia2022progcl,xia2022simgrace,xia2022survey,yue2022survey,liu2023dink} on graph-structured data are promising for molecular graph data~\cite{hu2020strategies,xia2023mole-bert,xia2023systematic,gao2022cosp}, just like the overwhelming success of pre-trained language models in natural language processing community~\cite{devlin2019bert,zheng2022using}.

The 3D molecular graph is composed of nodes (atoms), and their positions in 3D space and edges (bonds). The advantage of using 3D geometry is that the conformer information is critical to many molecular properties, especially quantum properties. In addition, it is also possible to directly leverage stereochemistry information such as chirality given the 3D geometries. Recently, multiple works~\cite{schuett2017schnet,satorras2021n,du2022se,liu2022spherical,atz2021geometric} have developed message-passing mechanisms tailored for 3D geometries, which enable the learned molecular representations to follow certain physical symmetries, such as equivariance to translations and rotations. However, the calculation cost, alignment invariance, uncertainty in conformation generation, and unavailable conformations of target molecules limited the applicability of these models in practice.

\section{Discussion and Conclusion}
In this paper, we perform a comprehensive benchmark of representative models on molecular property prediction. Our results reveal that traditional machine learning models, especially tree models, can easily outperform well-designed deep models in most cases. These phenomena can be attributed to the unique patterns of molecular data and different inductive biases of various models. Specifically, the target function mapping molecules to properties are non-smooth, and some small changes can incur significant property variance. Deep models struggle to learn such patterns. Additionally, molecular features carry meanings individually and deep models would undesirably mix different dimensions of molecular features. Our study leaves an open question for future research: Can our findings and methods be generalized to other AIDD tasks including drug-target interactions (DTIs) prediction~\cite{ozturk2018deepdta,xia2022towards}, drug-drug interactions (DDIs) prediction~\cite{li2021an}, and protein representation learning~\cite{hu2022protein,tan2023global}?

\end{document}